# A Hybrid Surrogate for Electric Vehicle Parameter Estimation and Power Consumption via Physics-Informed Neural Operators

Hansol Lim[1,2], Jongseong Brad Choi[1,2*], Jee Won Lee[1,2], Haeseong Jeoung[3], and Minkyu Han[3]

[1]Department of Mechanical Engineering, State University of New York, Korea, 21985, Republic of Korea
[2]Department of Mechanical Engineering, Stony Brook University, Stony Brook, NY 11794, USA
[3]Division of Advanced Vehicle Platform, Hyundai Motor Company, 13529, Republic of Korea
*corresponding author

*Abstract*—We present a hybrid surrogate model for electric vehicle parameter estimation and power consumption. We combine our novel architecture Spectral Parameter Operator built on a Fourier Neural Operator backbone for global context and a differentiable physics module in the forward pass. From speed and acceleration alone, it outputs time-varying motor and regenerative-braking efficiencies, as well as aerodynamic drag, rolling resistance, effective mass, and auxiliary power. These parameters drive a physics-embedded estimate of battery power, eliminating any separate physics-residual loss. The modular design lets representations converge to physically meaningful parameters that reflect the current state and condition of the vehicle. We evaluate on real-world logs from a Tesla Model 3, Tesla Model S, and the Kia EV9. The surrogate achieves a mean absolute error of 0.2 kW (≈1% of average traction power at highway speeds) for Tesla vehicles and ≈0.8 kW on the Kia EV9. The framework is interpretable, and it generalizes well to unseen conditions, and sampling rates, making it practical for path optimization, eco-routing, on-board diagnostics, and prognostics health management.

## 1. Introduction

Accurate battery consumption surrogates are crucial for trip planning, eco-routing, and on-board diagnostics as EV adoption grows [1]. Classical power-based vehicle models decompose tractive effort into aerodynamic drag, rolling resistance, grade, and inertia. When well calibrated, these models are lightweight and interpretable for on-board use [2]. In practice, however, calibration can drift due to factors such as tire wear, payload variations, auxiliary system usage, and climate conditions. This motivates the development of models that adapt to a vehicle's current state rather than relying on static factory settings [2, 3].

Prior work spans a spectrum from white-box physics to black-box machine learning methods. Physics-based approaches are transparent but sensitive to parameter error; purely data-driven models capture nonlinear patterns yet can be data-hungry, unstable in unseen conditions, and limited in interpretability for failure modes [4–6, 8]. Recent surveys emphasize this tradeoff and the need for methods that keep physics in the loop while adapting to real-world variability [4].

However, there are growing demands for hybrid approaches in the field to combine the strengths of both data-driven methods and physics. A prominent approach in this domain is the use of Physics-Informed Neural Networks (PINNs) [7, 26, 27]. They embed governing equations directly into the loss and have shown promising results for inverse problems, but they can suffer from stiff loss landscapes, sensitivity to weighting between data and physics terms, and optimization, especially on noisy, real-world signals [7, 8]. These issues limit generalizability across different sensors, operation environments, and vehicle conditions without careful re-tuning.

A recent trend in deep learning is operator learning [10]. It addresses several of these limits by learning maps between function spaces rather than pointwise regressors. The Fourier Neural Operator (FNO) captures global interactions efficiently in the spectral domain and is largely discretization-invariant [9]; Physics-Informed Neural Operators (PINO) further fuse physical constraints into the operator training for better generalization on parameterized dynamics [11]. These properties make neural operators attractive for EV energy modeling, where sampling rates between sensors and trip statistics vary.

Building on these ideas, we introduce EV-PINO, a Hybrid Surrogate for Electric Vehicle Parameter Estimation and Power Consumption via Physics-Informed Neural Operators. It utilizes a novel architecture Spectral Parameter Operator inspired by PINO and a fully differentiable EV physics module that is embedded directly into the network. From vehicle speed and acceleration log data, it outputs time-varying, bounded physical parameters (effective mass, drag, rolling resistance, motor efficiency, regenerative efficiency, and auxiliary power), and estimates vehicle power consumption. The model reflects the vehicle's current condition, operating environment, and because of embedded differentiable physics module, the surrogate

[1,2]Hansol Lim, [1,2]Jongseong Brad Choi, and [1,2]Jee Won Lee are with [1]Department of Mechanical Engineering, State University of New York, Korea, 21985, Republic of Korea, [2]Department of Mechanical Engineering, Stony Brook University, Stony Brook, NY 11794, USA, (e-mail: hansol.lim@stonybrook.edu; jongseong.choi@stonybrook.edu; jeewon.lee@stonybrook.edu). [3]Haeseong Jeoung, and [3]Minkyu Han are with [3]Division of Advanced Vehicle Platform, Hyundai Motors Company, 13529, Republic of Korea, (e-mail: hsjeoung@hyundai.com; minkyu.han@hyundai.com).
This work is supported by the National Research Foundation of Korea (NRF) grant funded by the Korea government (MSIT)–(No. RS-2025-05515607) and Hyundai Motor Company. (Corresponding author: Jongseong Brad Choi).

clearly distinguishes vehicle states (constant velocity, acceleration, deceleration) in its representations without manually labelling them.

*Contributions*
1. We developed a novel surrogate, EV-PINO, for EV parameter estimation and power consumption. EV-PINO maps vehicle speed and acceleration data to battery power by estimating parameters such as motor efficiency, regenerative braking efficiency, auxiliary power, coefficient of aerodynamic drag, coefficient of rolling resistance, and effective mass. It can robustly identify these parameters in a modular system that is fully interpretable and expandable.
2. The surrogate has two modules: (i) Spectral Parameter Operator, which is an FNO-backboned EV Parameter estimator with lightweight heads that focus on distinct, smaller representations of power terms. When aggregated, it outputs a full, expressive representation of the whole EV dynamics. This is possible due to its pair, (ii) Differentiable Physics Module. We removed the typical physics residual loss used in PINNs and embedded the dynamics equations directly in the forward pass. This hard-constrains the network's search space within physics hypothesis space. When synergized with Spectral Parameter Operator module which supports global data contexts, it helps the network quickly converge into physically consistent solutions.
3. We trained and validated our framework on real-world data from Tesla Model 3, Model S, and Kia EV9. It achieved low MSE and RMSE and the learned parameters converge to realistic values that reflect each vehicle's current condition. We compared these values to factory specs, and previous methods.
4. It can detect shifts in parameters through time which may indicate aging, load changes, or HVAC usage without additional sensors or inputs. This can be useful for applications in on-board monitoring, eco-routing, battery optimizations, and prognostics health management.

## 2. RELATED WORK

### 2.1. Modeling Strategies

EV power and energy modeling spans physics-driven ('white-box') and data-driven ('black-box') approaches. Hybrid ('gray-box') models combine the strengths of both. White-box approaches begin from vehicle modeling from the first-principles. This approach could be from dynamics of the vehicle, or from battery electrochemical and thermal formulations [4, 13, 15]. Their main advantage is interpretability, robustness across system platforms. The governing physics acting on the vehicle is invariant, while only the parameters differ across vehicles. This enables engineers to investigate failure modes and analyze the method systematically. However, accuracy hinges on calibrated parameters that drift with tire wear, payload, climate, and battery aging unless these values are re-identified in-situ. Recent applied studies emphasize how auxiliary systems (HVAC, lighting, infotainment) and environmental contexts inject non-trivial load variance that simple physics baselines often misattribute if not modeled explicitly [1, 2, 4].

On the other side, black-box methods learn mappings from signals (speed, acceleration, temperature, current, voltage histories) to power using regressors and deep neural networks (DNNs). Surveys show that black-box regressions can deliver useful range prediction within fixed operating regimes, but generalization to varying conditions is limited without explicit physics or careful feature design [1]. In battery state estimation, deep learning has had a clear impact. DNNs and sequence models such as Long Short-Term Memory (LSTM) networks improve state-of-charge (SOC) tracking and short-horizon prediction by capturing nonlinear effects and driver patterns [13–17]. Hybrid deep learning model variants like Particle Swarm Optimization (PSO) based tuning networks, cascaded learning networks further reduce variance and improve robustness on real trip data [15, 16]. Yet pure black-box models remain data-hungry, can extrapolate poorly under unseen sampling rates, vehicle conditions, operation environment, and offer limited physical interpretability for failure cases [19].

Not all black-box baselines are neural networks [12, 13]. LogPath is a log data only regression that identifies EV parameters by fitting a physics balance to recorded trajectories [20]. It estimates aerodynamic drag, rolling resistance, effective mass, and a single motor efficiency via least squares. We include LogPath as a transparent, code-available baseline in comparison to our work. It is fast and interpretable, but it does not capture time-varying efficiencies or condition drift.

Gray-box methods aim to keep the interpretability of physics while using machine learning to absorb unmodeled effects that cannot be easily predicted in real driving scenarios. [14, 15, 19].

A growing trend in gray-box method is physics-informed learning. Physics-Informed Neural Networks are regularized with physics residual. Our prior work, EV-PINN adopts this method. The model architecture embeds physics terms so that the network primarily learns physically consistent representations that satisfy the data. Recent papers and surveys describe growing interest in this direction because it reduces data demands and improves generalizability across hardware [1, 5, 19, 22, 26, 27].

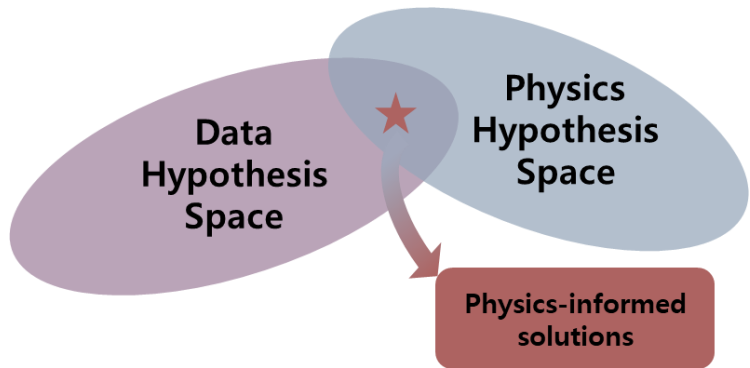

**Fig. 1.** High-level view of physics-informed learning: data narrows candidate functions while embedded physics constrains solutions to those consistent with EV dynamics.

The main goal of physics-informed learning is generalization under changing conditions, environments, and domain shifts while retaining an interpretable path back to physical quantities. [4, 6, 7]. Across these methods, we can define some important factors in gray-box surrogates.

Across the gray-box methods, several key designs consistently emerge. First is parameter drift and interpretability. While physics-based baselines are transparent, they require heavy, continuous computation to remain adaptive to changing conditions. This challenge is compounded by issues of resolution and context, as pointwise regressors trained at a single sampling rate often fail under new or unseen conditions. We need a new approach that can operate within different resolution and platforms [1,2, 7].

### 2.2. Weaknesses of Physics-Informed Neural Networks

Our prior work, EV-PINN [18] is a Physics-Informed Neural Network that was trained on real world driving data. It demonstrated that embedding EV dynamics into a neural network can accurately estimate instantaneous battery power and recover physically meaningful parameters directly from vehicle log data. It enabled inverse identification of system parameters without reliance on manufacturer calibration.

However, PINNs can be sensitive to loss weighting and initialization on noisy, real-world signals. EV-PINN had limitations from the inherent nature of the PINN architecture that motivated us for a different, reworked approach:

- **Instance-tied function approximation.** PINNs learn one solution rather than a mapping between function spaces. This makes the model data-specific. PINN models typically required heavy adjustment to hyperparameters then retraining when faced against different conditions of same system [8, 14, 15].
- **Limited global context.** Classical PINNs have MLP backbone and suffer from a limited global context. This point-wise processing makes them computationally inefficient when evaluating entire data profiles. While incorporating a Convolutional Neural Network (CNN) can introduce a local receptive field to aggregate information within a fixed window, it still fails to capture the broader, global context of the full system.
- **Resolution dependence.** Standard PINN implementations are trained at a fixed sampling rate; performance degrades or completely fails when the model faces samples at different resolution [11, 19]. For example, our prior work on using PINN on EV power prediction had issues regarding generalization across different resolutions. [14, 15, 18].
- **Training instabilities.** PINNs suffer during optimization process. They have stiff loss landscapes, imbalance between data/physics terms, and sensitivity to initialization—that makes convergence challenging in complex dynamics [14, 15].

These observations point toward operator learning. Neural Operators learn maps between function spaces and are designed to be resolution invariant. It allows inference on meshes or sample rates unseen during training [11]. Among Neural Operators, the Fourier Neural Operator is promising architecture. It captures global interactions by learning within the spectral domain and has shown strong accuracy and efficiency on parametric dynamics. Extending this idea, Physics-Informed Neural Operators (PINO) incorporate governing equations directly into operator learning, often generalizing where classical PINNs struggle to converge [9, 10, 11, 25].

## 3. PRELIMINARIES

*Operator Learning*

Let $X$ and $Y$ be Banach (typically Hilbert) spaces of functions on a domain $D \subset \mathbb{R}^d$. Define $X = L^2(D; \mathbb{R}^{d_a})$ and $Y = L^2(D; \mathbb{R}^{d_u})$, where d is the spatial dimension and $d_a$, $d_u$ are the input, output channel counts, respectively. Many physical systems define an operator $\mathcal{G}: X \to Y$ that maps an input field $a(x)$ to an output field $u(x)$. Given training pairs $\{(a_i, u_i)\}_{i=1}^N$, operator learning seeks a parametrized surrogate $\mathcal{G}_\theta: X \to Y$ minimizing empirical risk [25],

$$\min_\theta \frac{1}{N} \sum_{i=1}^N \mathcal{L}(\mathcal{G}_\theta(a_i), u_i). \tag{1}$$

Unlike pointwise regressions, $\mathcal{G}_\theta$ is defined on function spaces and can be evaluated on new discretizations of $D$.

*Neural Operators*

A neural operator composes layers that act on feature fields $v_\ell: D \to \mathbb{R}^{C_\ell}$:

$$(\mathcal{K}_\theta v_\ell)(x) = \int_D \kappa_\theta(x, y, a(y)) v_\ell(y) dy, \tag{2}$$

$$where\ v_{\ell+1}(x) = \sigma(\mathcal{K}_\theta v_\ell(x) + W_\ell v_\ell(x)) \tag{3}$$

with a learnable kernel $\kappa_\theta$, pointwise map $W_\ell$, and nonlinearity $\sigma$. Because layers are defined on function spaces, the trained $\mathcal{G}_\theta$ is resolution invariant [10].

*Fourier Neural Operators*

The Fourier Neural Operator (FNO) realizes $\mathcal{K}_\theta$ by spectral convolution[9]. Let $\hat{v}_\ell = \mathcal{F}(v_\ell)$ and let $P_m$ keep the lowest $m$ Fourier modes. Then,

$$v_{\ell+1}(x) = \sigma(W_\ell v_\ell(x) + \mathcal{F}^{-1}(R_\ell(P_m(\hat{v}_\ell)))) \tag{4}$$

where $W_\ell$ is a $1 \times 1$ linear mixing across channels. For each retained mode $k$, $R_\ell(k)$ is a learned complex matrix of shape $C_{out} \times C_{in}$ that mixes channels in the Fourier domain. $\mathcal{F}$ and $\mathcal{F}^{-1}$ are the 1-D Fourier transform and its inverse along the sequence (data axis). We apply $R_\ell$ only to the lowest $m$ positive frequencies and enforce conjugate symmetry so the output is real. This yields a global receptive field with per-layer cost $O(n \log n)$ for sequences of length $n$.

There are some useful properties that arise from FNO:

- **Function-to-function mapping.** As a neural operator, FNO learns a mapping between infinite-dimensional function spaces. This is a fundamental distinction from standard neural networks which learn mappings between finite-dimensional spaces.
- **Global receptive field and efficiency.** Convolution by low modes produces global interactions with $O(n \log n)$ per-layer complexity (through FFTs).
- **Resolution-invariance.** The spectral kernel defines a resolution-invariant map. Training and inference can occur at different resolutions, unlike standard CNN, RNN, PINN input-to-output mapping.
- **Inductive bias.** Truncating to dominant modes regularizes small-scale noise while retaining large-scale physics trends. This helps improvements in generalization of PDE/ODE families as shown in the original paper.

*Physics-Informed Neural Operators*

Physics-Informed Neural Operators augment neural operators with physics residual penalties (like PINN) using automatic differentiation to evaluate differential constraints at collocation points [11]. For a governing relation $\mathcal{N}(u; a) = f$ on a domain $D$ with boundary, initial terms $\mathcal{B}(u) = g$ on $\partial D$, we train $\mathcal{G}_\theta$ to minimize

$$\mathcal{L}(\theta) = \frac{1}{N} \sum_{i=1}^N \|u_\theta(a_i) - u_i\|_2^2 + \lambda_r \frac{1}{|X_r|} \sum_{x \in X_r} \|\mathcal{N}(u_\theta; a)(x) - f(x)\|_2^2 + \frac{1}{|X_b|} \sum_{x \in X_b} \|\|\mathcal{B}(u_\theta)(x) - g(x)\|\|_2^2 \tag{5}$$

which has been shown to improve robustness and zero-shot generalization compared with purely data-driven operators and pointwise regressor PINNs.

## 4. IMPLEMENTATION

### 4.1. System Overview

In this section we provide implementation details on EV-PINO. All implementations are done in Ubuntu 24.04 LTS, PyTorch 2.5.1 and CUDA 12.4 on NVIDIA RTX 5090 32GB GPU. For further details, please check our GitHub repository for the code: *https://github.com/SUNY-MEC-MEIC-Lab/EV-PINO*

We have designed a PINO-inspired hybrid surrogate that is divided in two stages:
1. **Spectral Parameter Operator**. A neural operator built upon on an FNO backbone. It uses spectral layers to efficiently capture global context from vehicle speed and acceleration. Its purpose is to understand and model EV dynamics by finding effective physical parameters that satisfy both data and physics.
2. **Differentiable Physics Module.** Instead of using a physics residual loss, we embed EV dynamics directly in the forward pass. Because the network has global context, we modularize the heads to specialize in learning distinct aspects of the data. With global context from our Spectral Parameter Operator, the module helps network learn representations that align with physics.

To ensure stability and physical realism, we constrain the model by bounding outputs of the Spectral Parameter Operator to realistic EV operation ranges. Also, the embedded differentiable physics module acts as a fail-safe core. Even under sudden spikes or outliers, the model's output structure is always constrained by the underlying physics to prevent failures and attributing deviations to specific terms so the balance among terms is preserved.

### EV Physics Model

For a longitudinal EV on flat roads, the standard tractive-effort balance decomposes wheel-side mechanical power into aerodynamic drag, rolling resistance, and inertial terms:

$$P_m(t) = \frac{1}{2}\rho A C_d v^3 + C_{rr}mgv + mav \tag{6}$$

with speed $v(t)$, acceleration $a(t) = \dot{v}$, air density $\rho$, frontal area $A$, drag $C_d$, rolling $C_{rr}$, mass $m$, and gravity g.
For this study, we exclude the gravitational force component ($F_{gravity} = mgv\sin\theta$) by assuming predominantly flat terrain. We combine $P_m$ motor efficiency with $\eta$, regenerative braking efficiency $\mu$ (when $a < 0$), and auxiliary power $P_{aux}$ from HVAC, lights, and accessories,

$$P_{bat}(t) = \frac{1}{\eta}P_m + \mu P_{m,a<0} + P_{aux}. \tag{7}$$

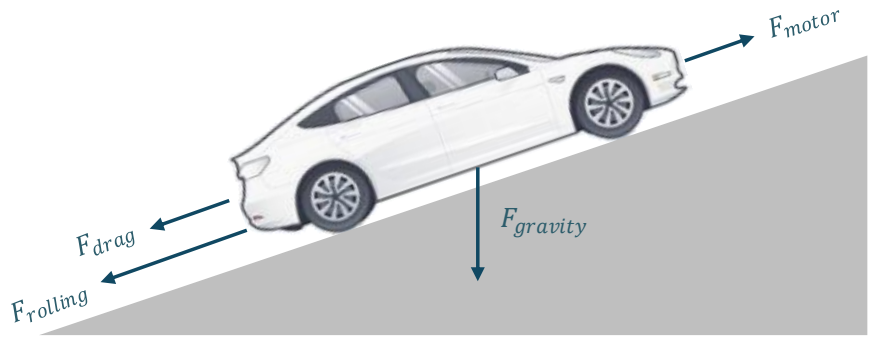

**Fig. 2.** Electric-vehicle free-body diagram and power balance showing traction, aerodynamic drag, rolling resistance, and grade forces during motion.

These modeling choices follow established vehicle dynamics from free-body diagram. In our previous work [18], regenerative braking efficiency $\mu$ was prone to be absorbed by total motor efficiency, $\eta$ because, during deceleration, it was not easy for the network to differentiate the balance between actual power and the regenerated power. Hence, we have changed the regenerative braking efficiency to be a separated term from motor efficiency.

### EV Data Collection and Augmentation

For the experimental validation, we collected real-world data using OBD-II scanner (OBDLink MX+ model) on Tesla Model 3 Long Range and Tesla Model S. We were also provided with Kia EV9 Long Range 4WD log data from Hyundai Motors for the validation of the framework.

The log data consisted of time-series vehicle speed data and battery pack current and voltage. From our previous work, we learned that many of the instabilities in training come from noise in the data. Hence for our new work, we apply Savitzky-Golay (SG) filtering [28] which will help the network focus on real vehicle dynamics rather than measurement artifacts:

$$v_k = \sum_{j=m}^{m} c_j^{(0)} v_{k+j} \tag{8}$$

where $v$ is the log data speed and coefficients $c_j^{(0)}$ are the least-squares weights for the zeroth derivative of the order-p polynomial.

After applying the filter, we can compute acceleration from the analytic first derivative of the fitted SG polynomial. This yielded much smoother acceleration compared to computing acceleration using finite differences from raw data. We chose SG polynomial order 3 and a window length of 1.1 second.

*Input Normalization*

We form input windows of length $L$ with stride $s$ over the logged sequence, using $[v, a]$ as inputs and battery power $P_{bat}$ as the target. Speed and acceleration are standardized (z-score) with a StandardScaler fitted on the training split. Physics is always computed on unnormalized $[v, a]$ by inverting the scaler at each batch. Each window is augmented with a normalized positional grid $\xi \in [0,1]$ to provide absolute phase to the operator.

### 4.2. Spectral Parameter Operator

We train a Spectral Parameter Operator that acts as state estimator feeding a differentiable physics module:

$$\Theta_\theta : \mathcal{H}(v, a) \mapsto (C_d, C_{rr}, m, P_{aux}, \eta, \mu). \tag{9}$$

We can structure the network so that it reflects the EV differential equation directly (eq. 7):

$$P_{bat}(t) = \frac{ReLU(P_m(t))}{\eta} + \mu \, ReLU(-P_m(t)_{a<0}) + P_{aux}. \tag{10}$$

We chose to embed the physics directly into the forward pass rather than having a compute-heavy physics residual loss. We were inspired by papers that have modular designs such as Deep Lagrangian Networks [24], Transformers [29] (which is also a special case of Neural Operator), and Mixture-of-Expert system [30]. These studies have shown that smaller subnetworks (heads) can focus on different, smaller representations then later aggregate them to have richer, full representations. This upgrade also helps interpretability, and reduces the search space. We found that this modular structure fits operator learning especially well. A single spectral forward pass gives global context. The heads learn parameters that govern each force term rather than learning the whole physics at once.

*Fourier Neural Operator Backbone*

We implement $\Theta_\theta$ with an FNO backbone with 1-D Fourier Neural Operator with $L_{spec}$ spectral blocks, $W$ channels, and $m$ retained Fourier modes per layer. The lift MLP maps $[v, a, \xi] \in \mathbb{R}^{B \times W \times 3}$ to width $\mathbb{R}^{B \times W \times L}$ [21]:

$$h_0 = Linear \circ GELU \circ Linear([v, a, \xi]) \in \mathbb{R}^{B \times W \times L}. \tag{11}$$

We stack $L_{spec}$ spectral blocks where each block $\ell$ computes a residual update:

$$x_{\ell+1} = x_\ell + GELU\big(SPEC_\ell(x_\ell) + MLP_\ell(x_\ell)\big) \tag{12}$$

where the spectral path operates along the sequence axis via truncated rFFT,

$$S_\ell(x) = \mathcal{F}^{-1}\big(W_\ell \odot P_m(\mathcal{F}(x))\big) \tag{13}$$

where $P_m$ keeps the lowest m modes, and $W_\ell$ are learned complex weights. The pointwise mixes channels only:

$$MLP_\ell(x) = W_{\ell,2} GELU(W_{\ell,1} x). \tag{14}$$

After $L_{spec}$ blocks we obtain feature map $Z = x_{L_{spec}} \in \mathbb{R}^{B \times W \times L}$.

This residual form keeps the map near identity at init and stabilizes deep stacks. The Fourier path provides global receptive fields in $O(L \log L)$. The mode truncation makes the operator tolerant to changes in sampling rate and window length.

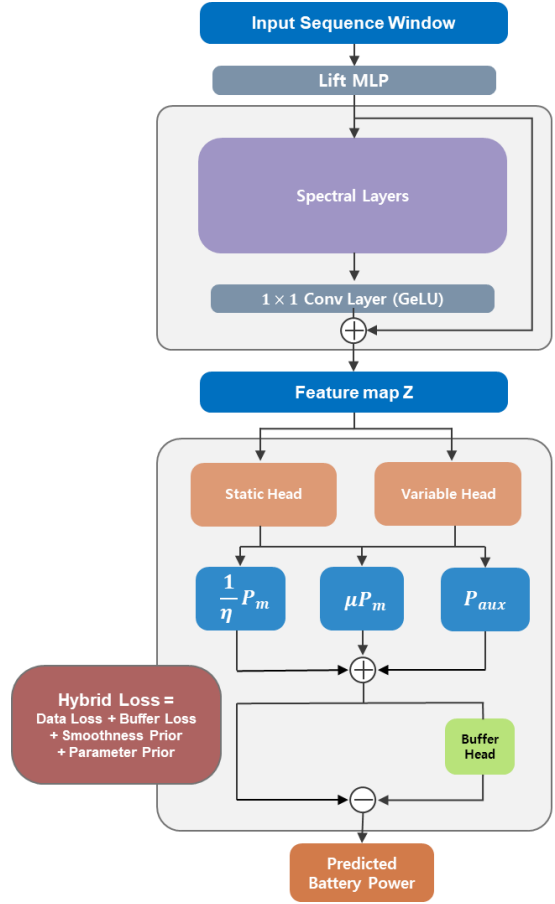

**Fig. 3.** Architecture block diagram. FNO trunk with lightweight heads producing parameters, followed by an embedded physics layer to compute power.

*Parameter Bounds*

We estimate time-varying parameters and scalar parameters through variable head and static head. Heads used to estimate parameters output by learning baseline scalars $(C_d, C_{rr}, m, P_{aux}, \eta, \mu)$ with unconstrained parameters passed through a sigmoid to enforce realistic bounds (e.g., Tesla Model 3 $C_d = [0.20, 0.30]$). We can define bounds as:

$$Bound(x; a, b) = a + (b-a)\sigma(x), \quad \sigma(x) = \frac{1}{1+e^{-x}}. \tag{15}$$

We found that some parameters such as $\eta$ and $\mu$ cannot be explained by a single constant because they vary hugely through different vehicle speed and time. We therefore let the variable head produce offsets and gate them by speed:

$$(\Delta\eta, \Delta\mu) = \tanh\left(\frac{H_{var}(Z)}{T_H}\right) \in [-1,1]^2 \tag{16}$$

where $Z \in \mathbb{R}^{B \times W \times L}$ is the latent feature map from Spectral Parameter Operator, $H_{var}$ is a $1 \times 1$ conv head with two output channels, $T_H$ is head temperature hyperparameter that controls the smoothness of the offsets, and $\Delta$ indicates baseline offsets.

We limit quick efficiency changes at low speeds by gating the time-varying heads with a smooth factor,

$$w(v) = \sigma\left(\frac{v - V_0}{S_V}\right) \in [0,1] \tag{17}$$

where $w \approx 0$ at low speeds and $w \approx 1$ at high speeds, $V_0$ and $S_v$ are hyperparameters (we used $V_0 = 18 \, m/s \approx 65 \, km/h$) that represents transition speed and slope that controls how quickly gate opens, respectively. We assigned these hyperparameters to suppress low-speed oscillations and open variations at high speeds.

The time-varying parameters are formed by adding bounded, speed-gated perturbations to global baselines found in the warm-up,

$$\eta_t = clip(\eta_0 + SPAN_\eta [w \odot \Delta \eta], \quad [\eta_{min}, \eta_{max}] \tag{18-1}$$

$$\mu_t = clip(\mu_0 + SPAN_\mu [w \odot \Delta \mu], \quad [\mu_{min}, \mu_{max}] \tag{18-2}$$

where $\odot$ is element-wise product and $clip(x; a, b) = \min(\max((x, a), b))$.

For Kia EV9, we also set $P_{aux}$ using third channel in the same way to demonstrate extensibility of the framework. However, as results will show later, $P_{aux}$ has no relationship to vehicle speed and acceleration so network cannot learn relationships. We do it for the purpose of showing extensibility and failure case when we force the network to learn something that not is not feasible.

We also have a separate $1 \times 1$ residual buffer head that outputs $P_{res}(t) \in \mathbb{R}^{B \times L}$. It is initialized at zero, so the model starts from pure physics and only uses residuals if needed. The buffer head acts as a compensator, absorbing small unmodeled power terms not captured by the physics module. This reduces interference between heads during training phase. It was a problem in our prior work [18], where networks tended to "cheat" by explaining unmodeled effects with other parameters. Networks were observed to absorb these effects into less sensitive parameters like vehicle mass or motor efficiency.

### 4.3. Training

*Hybrid Objective Loss*

Given log data pairs $\{v, a, P\}$, we predict $\theta_{raw} = [C_d, C_{rr}, m, \eta, \mu, P_{aux}]$ and minimize a hybrid loss:

$$\mathcal{L}(\theta) = \underbrace{\frac{1}{BL} \sum_{b,t}^{BL} \|P_{data} - P_{pred}\|_2^2}_{\text{data loss}} + \underbrace{\lambda_{buff} \frac{1}{BL} \sum_{b,t}^{B} \|P_{residual}\|_2^2}_{\text{buffer loss}} + \underbrace{\lambda_{smooth} \frac{1}{B(L-1)} \sum_{b,t}^{B} \|\Delta \eta\|_F^2 + \|\Delta \mu\|_F^2}_{\text{smoothness prior}} + \underbrace{\lambda_{param} \|\theta_{raw}\|_2^2}_{\text{parameter prior}} \tag{19}$$

where $\Delta \eta$ and $\Delta \mu$ are first-order time differences. $r$ is a residual vector that contains residuals from all parameter heads, $F$ is Frobenius norm. Each component serves a specific purpose:
- **Data loss** term measures if the prediction aligns with data and anchors the model into data space.
- **Buffer loss** penalizes the residual buffer head to be small so that it only absorbs small, unmodeled effects. It prevents network from cheating by pushing everything into residual instead of looking for appropriate parameters.
- **Smoothness prior** discourages discontinuities. This ensures the time-varying parameters to evolve smoothly rather than having values jumping up and down.
- **Parameter prior** is a soft prior that regularizes baseline parameters back inside their physically plausible ranges to prevent unrealistic values. We use the raw values before the sigmoid bound to prevent mid-range bias.

*Two-Stage Training*

We divided the training into two stages:
1. **Physics-only warm up.** We freeze the operator and all parameter heads. We focus on learning only baselines scalars $(C_d^{(0)}, C_{rr}^{(0)}, m^{(0)}, P_{aux}^{(0)}, \eta^{(0)}, \mu^{(0)})$ that best fits our differentiable physics module. This anchors the plant to a realistic starting point and removes early competition between constant parameters such as $C_d, C_{rr}, m, P_{aux}$ and varying parameters $\eta, \mu$. If the bounds for the parameter are big, the warm-up epochs could be increased so that the we can give parameters more time to stabilize.
2. **Operator Learning.** After the warm-up, we unfreeze the Spectral Parameter Operator and parameter heads. We jointly refine baselines and time-varying parameters through Adam with base learning rate of $3e - 4$. We use cosine annealing and an early stopping on validation MSE and restore to the best checkpoint. We implemented early stopping if the model has converged enough. However, we noticed that it may require longer epochs if the log data duration is big.

By optimizing the hybrid loss in two stage training phases, the model learns representation that satisfy real driving data and the underlying vehicle dynamics.

## 5. RESULTS AND VALIDATION

Below are results of EV-PINO on Tesla Model 3 Long Range data. Vertical dashes indicate change from warm-up stage to PINO training stage. Results for Tesla Model S and Kia EV9 are in Appendix A.

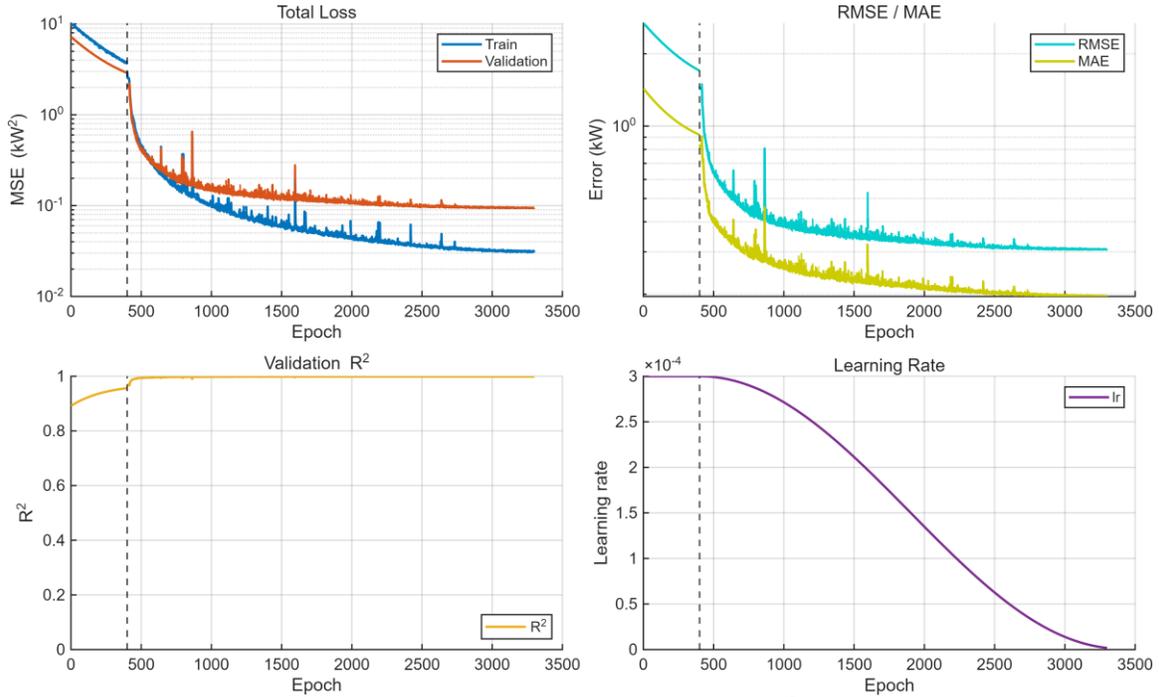

**Fig. 4.** Training curves for the Model 3 showing train and validation loss, $R^2$, and learning-rate schedule. Loss drops smoothly and the gap between train and validation stays small. Early stopping triggers near the best validation point. The warm-up and decay schedule yields stable optimization without spikes.

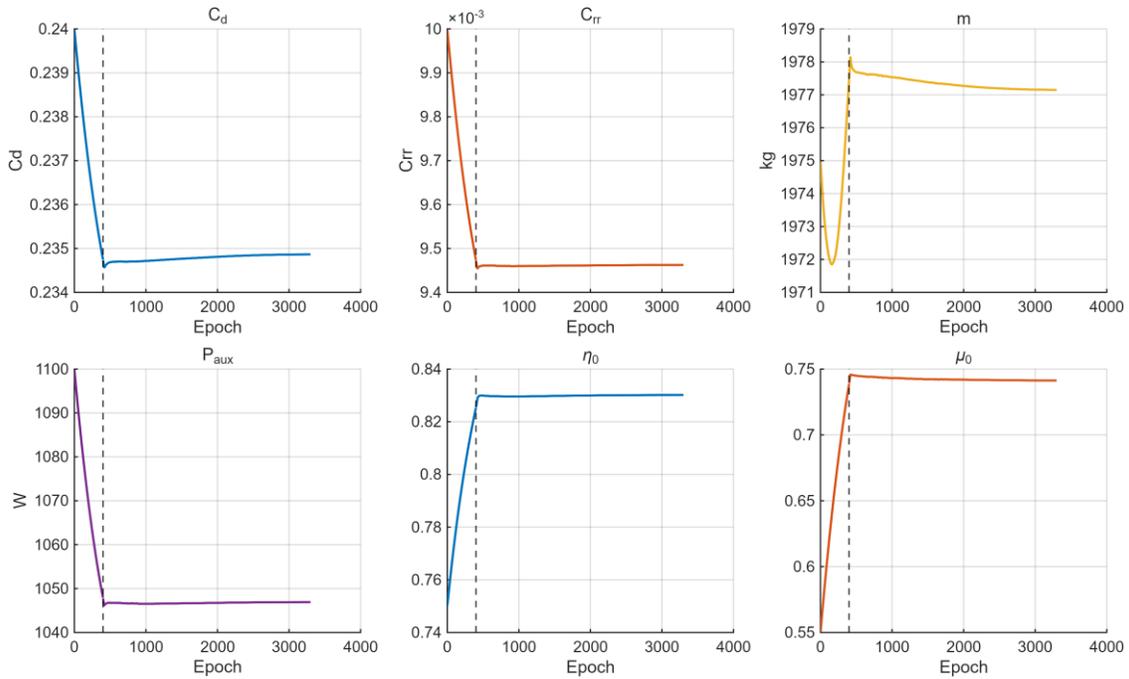

**Fig. 5.** Evolution of scalar parameters and efficiency heads over epochs. Starting from broad, generic priors, the estimates converge to vehicle-appropriate, physically plausible values within bounds. Trajectories flatten as uncertainty shrinks, and clipping becomes rare late in training. It shows that having smaller heads leads to stable convergence of parameters and reduces energy leaks between terms.

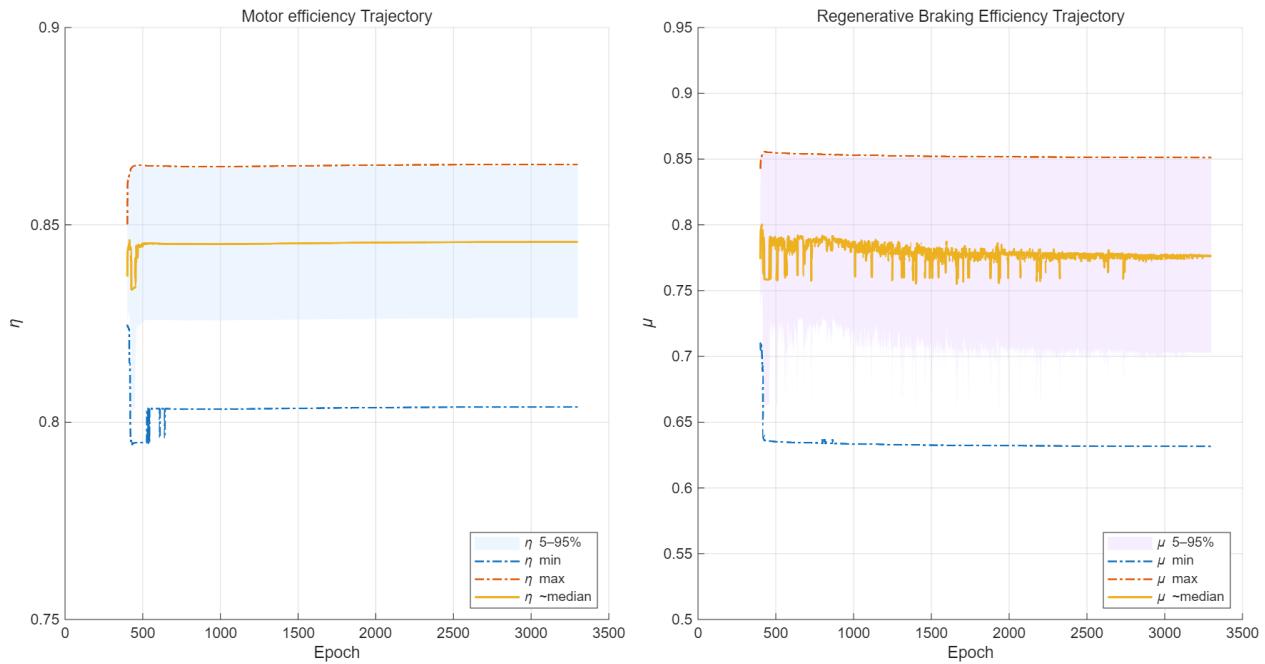

**Fig. 6.** Learned time-varying motor efficiency and regenerative-braking efficiency. Signals sharpen across epochs as noise is suppressed and trends align with operating regime. Motor efficiency rises under moderate load and drops at extremes, while regenerative braking efficiency has higher variance as it is affected hugely by driving style.

*Analysis on Results*

Tables 1 summarizes parameter identification and prediction performance. EV-PINO accurately recovers each vehicle's current-state physical parameters and, in turn, yields precise power estimates.

The learned mass captures curb weight plus driver and payload. Aerodynamic drag aligns with vehicle types (sedans (Model 3, S) $C_d = 0.2349, 0.2386$, SUV (Kia EV9) $C_d = 0.2788$).

| Parameter | Vehicle | Factory Specs | EV-PINO Estimates |
|---|---|---|---|
| $\eta$* | Tesla Model 3 LR | - | 0.8302 |
| | Tesla Model S | - | 0.8282 |
| | Kia EV9 LR 4WD | - | 0.8447 |
| $\mu$ | Tesla Model 3 LR | - | 0.7413 |
| | Tesla Model S | - | 0.7273 |
| | Kia EV9 LR 4WD | - | 0.7279 |
| $m$ | Tesla Model 3 LR | 1844 kg (curb weight) | 1977 kg |
| | Tesla Model S | 2170 kg (curb weight) | 2246 kg |
| | Kia EV9 LR 4WD | 2590 kg (curb weight) | 2784 kg |
| $C_{rr}$ | Tesla Model 3 LR | 0.0096 | 0.009462 |
| | Tesla Model S | 0.0096 | 0.01048 |
| | Kia EV9 LR 4WD | 0.00664 | 0.007214 |
| $C_d$ | Tesla Model 3 LR | 0.23 | 0.2349 |
| | Tesla Model S | 0.24 | 0.2386 |
| | Kia EV9 LR 4WD | 0.28 | 0.2788 |
| $P_{aux}$ | Tesla Model 3 LR | - | 1046 W |
| | Tesla Model S | - | 236 W |
| | Kia EV9 LR 4WD | - | 744 W |

**Table 1.** Learned Parameter Estimates. EV-PINO refines physically plausible initial guesses and converges to vehicle-specific, physically grounded values under the test conditions. *Electric vehicles are known to have ~90% motor efficiencies and may deteriorate with age.

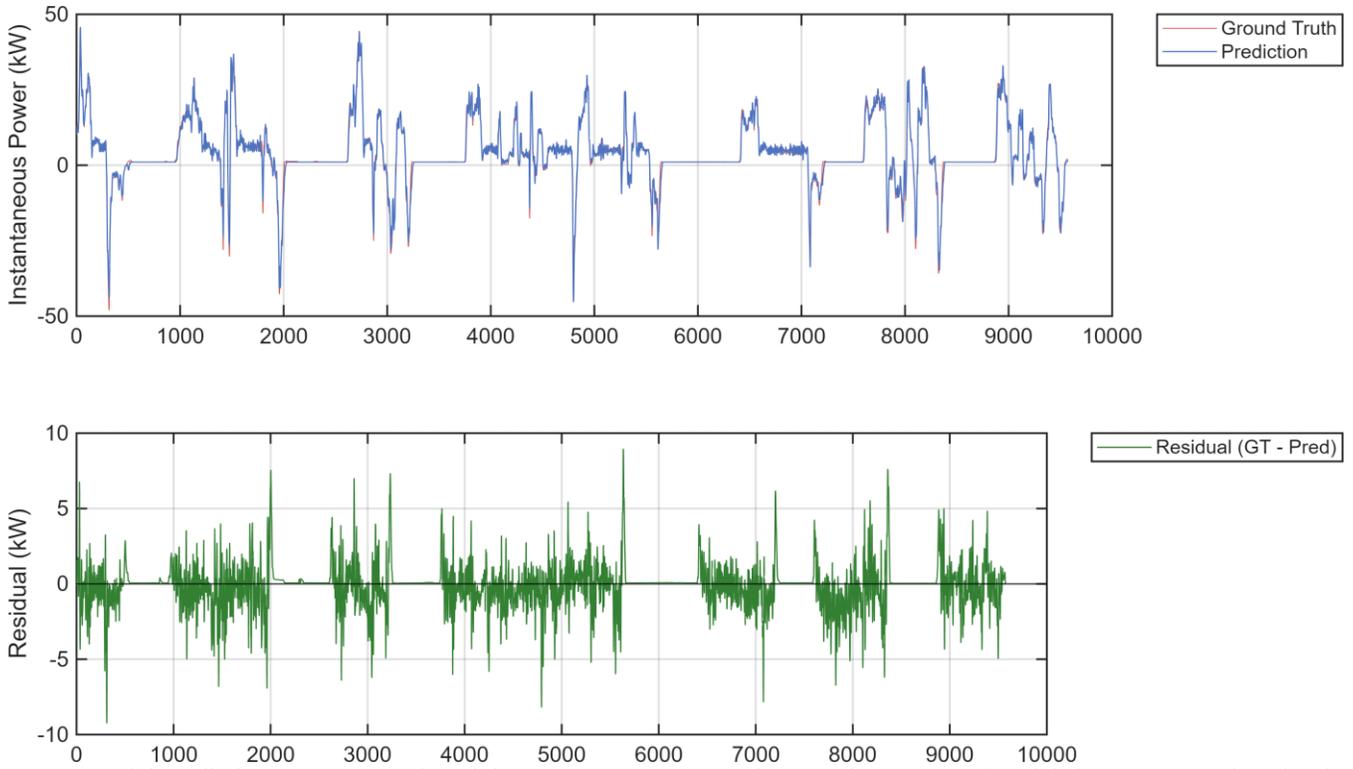

**Fig. 7.** Model prediction versus ground-truth battery power on a 15-minute cruise segment (Model 3). Curves overlap closely with small, near-zero-mean residuals and no visible phase lag. Rapid transients and steady-state plateaus are both captured, reflecting the benefit of global context plus embedded physics. The aggregate MAE and RMSE are 0.1968 kW 0.3078 kW, respectively.

| Metric | Tesla Model 3 LR | Tesla Model S | Kia EV9 LR 4WD |
|---|---|---|---|
| Total Loss ($kW^2$) | 0.031701542 | 0.100727499 | 0.284349207 |
| Validation Loss ($kW^2$) | 0.093798354 | 0.084527038 | 1.59140563 |
| MAE (kW) | 0.1968 | 0.1971 | 0.7835 |
| RMSE (kW) | 0.3078 | 0.2907 | 1.264 |
| Training Time (s) | 417.81 (Epoch 3000) | 76.68 (Early Stopping 1000) | 741.81 (Early Stopping 2750) |
| # Parameters | 690,697 | | 690,826 |

**Table 2.** Model performance comparison between different vehicles. All trained models show good metrics for a feasible surrogate. Kia EV9 have relatively higher metric values. This is due to the dataset range spanning roughly 5x more than Tesla datasets. Hence, they are similar when considering the scale of the dataset range.

Fig. 7 and Table 2 show model predictions for power consumption. For the Tesla vehicles, it gives very accurate MAE ≈ 0.2 kW and RMSE ≈ 0.3 kW considering that they peak up to ±50 kW (±250 kW for EV9).

The architecture remains small (~690k parameters). Training time varies with log complexity under early stopping (77s for Model S to >700s for EV9). Overall, EV-PINO learns a robust physical representation rather than merely fitting curves across vehicle state, condition, and environment.

| | Tesla Model 3 LR / S | Kia EV9 LR 4WD |
|---|---|---|
| **Lift MLP** | 33,920 | 33,920 |
| **Spectral Conv Layers** | 524,288 | 524,288 |
| **Per-layer 1x1 MLPs** | 132,096 | 132,096 |
| **Parameter heads** | 129 | 129 |
| **Variable heads** | 258 | 387 |
| **Scalar Baseline Neurons** | 6 | 6 |
| **Total # Parameters** | 690,697 | 690,826 |

**Table 3.** Model size and computational characteristics. Parameter counts are listed for EV-PINO variants under the same input window.

*Comparison Between Existing Works*

| Vehicle | Method | $\eta$ | $\mu$ | $m$ (kg) | $C_{rr}$ | $C_d$ | $P_{aux}$ (W) |
|---|---|---|---|---|---|---|---|
| Tesla Model 3 Long Range | Factory Specs | - | - | 1,844 | 0.0096 | 0.23 | - |
| | LogPath (Regression) | 0.8209 | - | 1,752 | 0.01313 | 0.01217 | 1,100 |
| | EV-PINN (MLP-PINN) | 0.7349 | 0.6743 | 1,975 | 0.00915 | 0.2349 | 1,100 (Hard-coded) |
| | **EV-PINO (FNO-PINO)** | **0.8302** | **0.7413** | **1,977** | **0.009462** | **0.2349** | **1046** |
| Tesla Model S | Factory Specs | - | - | 2,170 | 0.0096 | 0.24 | - |
| | LogPath (Regression) | 0.7248 | - | 2,145 | 0.0156 | 0.2877 | 390 |
| | EV-PINN (MLP-PINN) | 0.7248 | 0.7038 | 2,313 | 0.0110 | 0.2457 | 390 (Hard-coded) |
| | **EV-PINO (FNO-PINO)** | **0.8282** | **0.7273** | **2,246** | **0.01048** | **0.2386** | **236** |
| Kia EV9 Long Range 4WD | Factory Specs | - | - | 2,590 | 0.00664 | 0.28 | - |
| | LogPath (Regression) | 0.7540 | - | 2,862 | 0.008892 | 0.2564 | 550 |
| | EV-PINN (MLP-PINN) | 0.7983 | 0.7567 | 2,631 | 0.005609 | 0.2780 | 550 (Hard-coded) |
| | **EV-PINO (FNO-PINO)** | **0.8448** | **0.7279** | **2,784** | **0.007214** | **0.2788** | **744** |

**Table 4.** Comparison with specifications and prior methods (considering the code availability). Reported values are effective parameters identified from the given logs; they reflect vehicle, payload, grade, ambient conditions, and our physics parameterization. Manufacturer specs are nominal and trim-dependent, and prior methods fix or regularize different terms. Because the underlying coefficients are not directly observable in this setup, we use this table for plausibility, not for adjudicating a single "true" set of values.

| | Neural Networks | LogPath | EV-PINN | **EV-PINO** |
|---|---|---|---|---|
| Can model dynamics | ✓ | ✓ | ✓ | ✓ |
| Learns underlying physics | x | x | ✓ | ✓ |
| Estimates physical parameters | x | ✓ | ✓ | ✓ |
| Estimates Auxiliary power | x | ✓ | x (needs to be hard-coded) | ✓ |
| Estimates Regenerative Braking | x | x (absorbed into motor efficiency) | ✓ | ✓ |
| Adaptive to different sensor resolution | x | ✓ | x | ✓ |
| Adaptive to changing conditions | ✓ (if sufficiently trained) | x (linear regression) | ✓ | ✓ |
| Robust to extreme conditions | x (may fail to generalize) | x (linear regression) | x (may fail to generalize) | ✓ **(built-in differentiable physics)** |

**Table 5.** Comparison table between similar works that estimate EV parameters and power consumption.

The median values for the key time-varying coefficients—motor efficiency ($\eta$), regenerative-braking efficiency ($\mu$)—were recovered by each method. Our method, EV-PINO columns demonstrate higher motor efficiency ($\eta \approx 0.83 - 0.85$) and a clear regenerative-braking efficiency ($\mu \approx 0.65$–$0.73$). This is because the physics-informed architecture:

1. Isolates aerodynamic drag ($C_d$), rolling resistance ($C_{rr}$), mass ($m$), and auxiliary power ($P_{aux}$), preventing these losses from being misinterpreted as poor drivetrain efficiency.
2. Learns a separate channel for regenerative braking, rather than combining deceleration energy into a single slope. Motor efficiency values are higher in our results compared to other methods. This is due to PINO captures fast transients with the spectral trunk, which means less unmodeled power is absorbed by the motor efficiency. Regenerative braking efficiency is also higher due to modification in physics model. In EV-PINN, regenerative braking efficiency was divided by motor efficiency. However, in EV-PINO, we model these terms separately.
3. In contrast, earlier linear-regression work LogPath lacked these mechanisms. It grouped auxiliary powers, and regenerative braking losses into the traction term. This resulted in a lower apparent efficiency at ~0.72 and failed to estimate the parameters accurately.

*Model Sizing and Extensibility*

Table 3 shows the parameter sizes for the two variants of our surrogate model, with the only difference being an additional output head in the Kia EV9 version to predict an auxiliary variable ($P_{aux}$).
This change is minor, adding only 129 parameters, and demonstrates the flexibility of EV-PINO's modular design. Because the output heads are separate from the backbone, both model variants share the same core architecture and computational complexity. This modularity allows us to add new outputs as needed without altering the model's foundation.

*Base Architecture Comparison*

We compare MAE between MLP based PINN, Transformer based PINN [23] against EV-PINO for power prediction.

| Architecture | PINN (MLP) | | PINN (Transformer) | | **PINO** | |
|---|---|---|---|---|---|---|
| Vehicle type | Model 3 LR | Model S | Model 3 LR | Model S | **Model 3 LR** | **Model S** |
| MAE (kW) | 0.732 | 1.080 | 0.386 | 0.0790 | **0.196** | **0.197** |
| rMAE | 0.1013 | 0.1470 | 0.0612 | 0.0135 | **0.0295** | **0.0172** |
| rRMSE | 0.1012 | 0.1441 | 0.0736 | 0.0175 | **0.0460** | **0.0253** |
| Training time | 5m57s | 11m19s | 25s | 1m35s | **6m57s** | **1m17s** |
| # of parameters | 33,414 | | 453,562 | | 690,697 | |
| Resolution invariance | x | | x | | ✓ | |
| Inference | Point-wise mapping | | Point-wise mapping | | **Maps entire data set** | |

**Table 6.** Comparison on different base network architectures on Tesla Model 3 Long Range and Model S. We compare PINN (MLP), PINN (Transformer), and EV-PINO using MAE, rMAE, rRMSE, training time, and parameter count. Relative errors use mean positive battery power as the scale; training times reflect early stopping.

On Tesla Model 3 LR, EV-PINO cuts error by half relative to the Transformer PINN baseline. On Model S, the Transformer is slightly better in MAE; the gap is ~0.12 kW, about 0.5% of average traction power. Across both cars, EV-PINO adds interpretable, physically grounded parameters and works across sampling rates without retuning. A Transformer could be made semi-resolution invariant with continuous time encodings and multi-rate training, but we kept the implementation simple.

Although, Transformers are known for their brute performance. PINO wins in-terms of practicability. The PINN baselines predict sample-by-sample at a fixed rate. EV-PINO, as an operator, maps an entire trajectory to its power trace in one pass. This gives lower latency, easy batching, and resolution-invariant inference, which is exactly what on-board path optimization and vehicle health monitoring systems require.

*Computational Complexity*

Our choice of a PINO architecture based on the Fourier Neural Operator (FNO) provides a significant advantage over Transformers in efficiency too. While both model types can capture global context, Transformers are data-hungry and they scale with a quadratic complexity of $O(n^2)$. In contrast, FNO-based models require less data and scale with a much more efficient quasilinear complexity of $O(n \log n)$. This is possible because the FNO's weights exist in Fourier space. This allows the backward pass to reuse intermediate buffers from the FFT during training.

*Power Spectral Density Test*

We compare power spectral density using Welch's method. The peaks align across frequencies, which indicates that the model has learned correct dynamics.

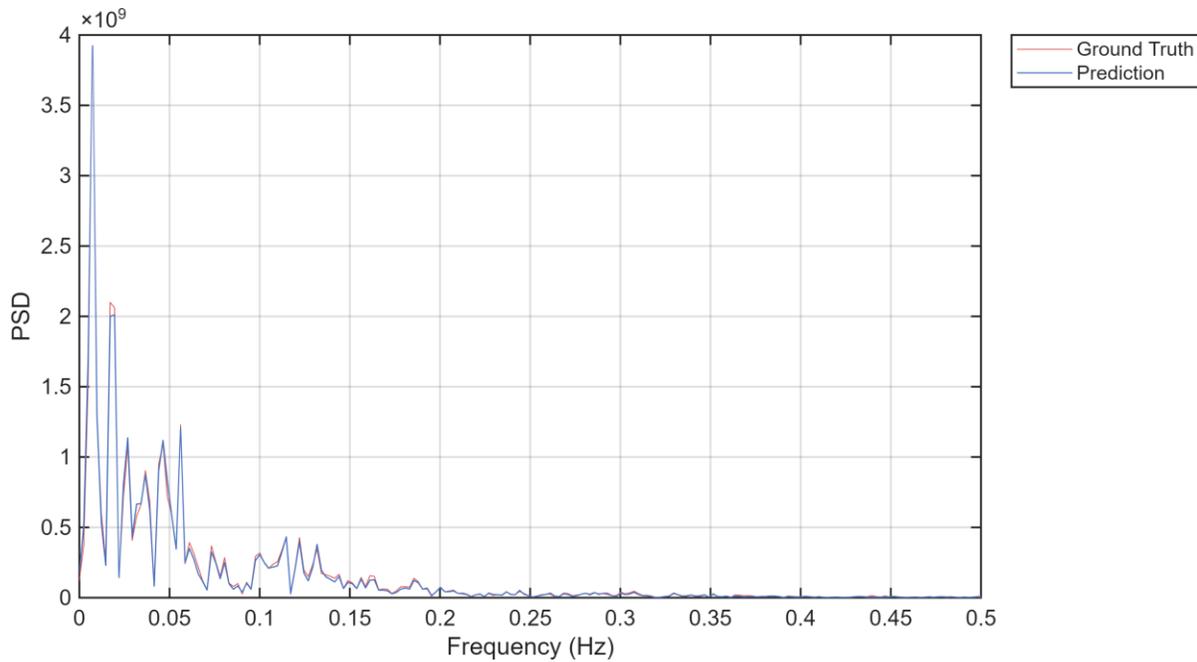

**Fig. 8.** Welch power spectral density of prediction and ground truth. Peaks align across low frequencies (drive-cycle components) and mid frequencies (acceleration transients), and the noise floor remains similar. The match indicates correct dynamic content without spurious harmonics. This frequency-domain agreement complements the low time-domain error.

## 6. CONCLUSION

In this work, we introduced EV-PINO, a novel hybrid surrogate architecture for electric vehicle parameter estimation and power consumption. Our framework successfully learned to identify key vehicle parameters—including effective mass, aerodynamic drag, rolling resistance, and time-varying motor and regenerative braking efficiencies—directly from real-world driving data.

Our evaluations on different EV platforms, including a Tesla Model 3, Tesla Model S, and Kia EV9, demonstrate that EV-PINO not only predicts battery power with high accuracy but also converges to physically meaningful parameter values. Most importantly, the model adapts to the vehicle's actual operating conditions, capturing variations from factors such as payload, component aging, and auxiliary system usage, which are not accounted for in static factory specifications.

The core strength of EV-PINO lies in its modular learning process using global context from the spectral domain and specialized heads that blend with a differentiable physics module embedded in the network, producing solutions aligned with each distinct power term. Spectral Parameter Operator focuses on learning complex, time-varying parameter dynamics from data, while the differentiable physics modules ensures that predictions remain grounded in established vehicle dynamics. This design makes the framework inherently robust and interpretable. EV-PINO provides a practical and reliable tool for a range of applications, from on-board energy management and eco-routing to long-term vehicle health monitoring. This framework also underscores the potential of hybrid, Physics-Informed Neural Operator models to create robust surrogates. We hope this work inspires other researchers extend the idea to complex engineering systems other than EVs such as gasoline cars, drones, robots, and much more.

*Credit Authorship Contribution Statement*

**Hansol Lim:** Conceptualization, Data curation, Formal analysis, Investigation, Methodology, Project administration, Software, Validation, Visualization, Writing – original draft, Writing – review & editing.

**Jongseong Brad Choi**: Conceptualization, Funding acquisition, Project administration, Resources, Supervision, Writing – review & editing.

**Jee Won Lee:** Data curation, Investigation, Resources, Validation, Visualization, Writing – review & editing

**Haeseong Jeoung:** Data curation, Project administration, Resources.

**Minkyu Han:** Data curation, Project administration, Resources.

*Data Availability*

All code, scripts, and configuration files, dataset used in this study (Vehicle log data) are publicly available at: https://github.com/SUNY-MEC-MEIC-Lab/EV-PINO. The repository contains the exact programs used to produce our results, with instructions to reproduce data shown on figures and tables.

*Acknowledgments*

This work is supported by the **National Research Foundation of Korea** (NRF) grant funded by the Korea government (MSIT) (No. RS-2025-05515607) and by **Hyundai Motor Company**.

APPENDIX A

This appendix provides graphical complements to support the main results for Tesla Model S, and Kia EV9 Long Range 4WD.

**Tesla Model S**

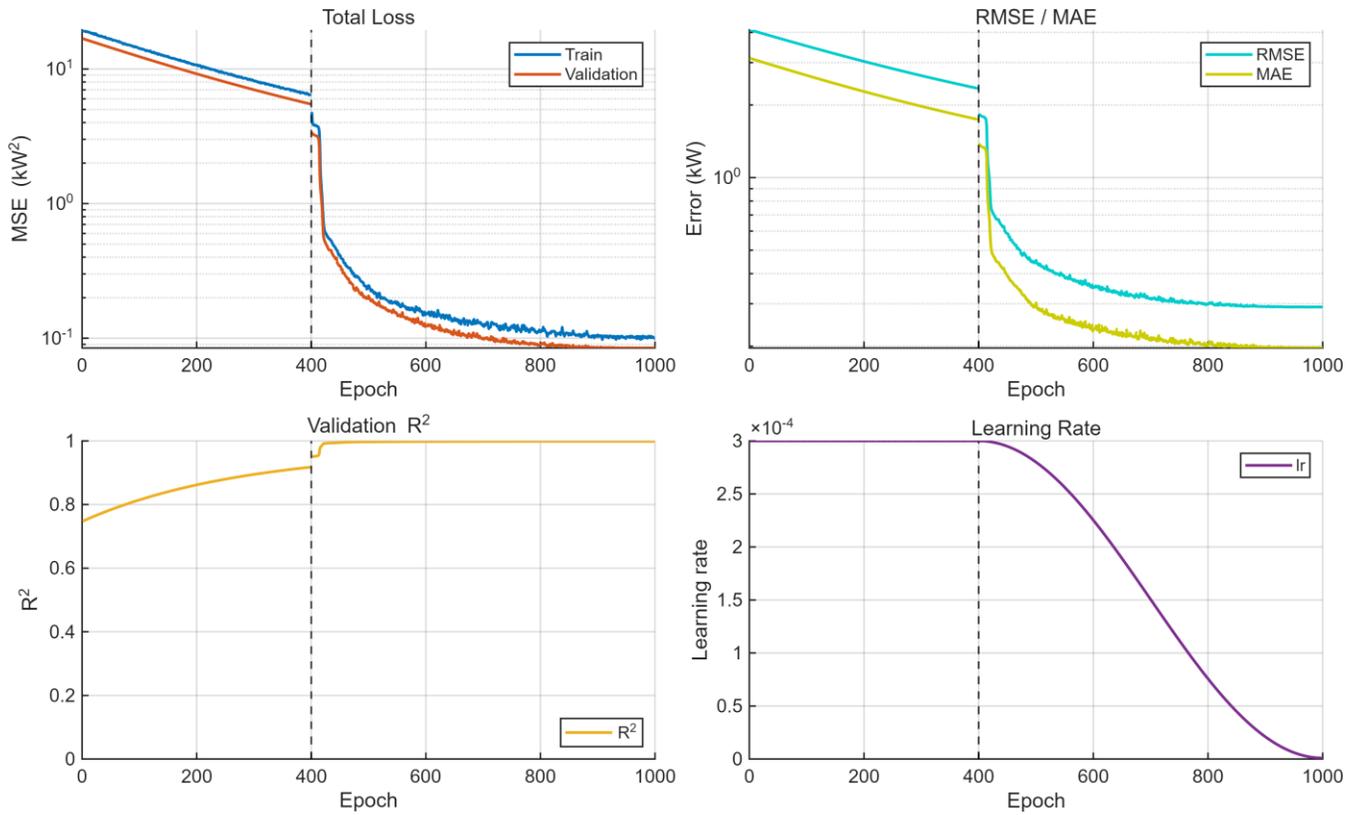

**Fig. A-1.** Model S training curves. Smooth loss, small training and validation loss gap, and early stopping at peak validation epoch (~1000).

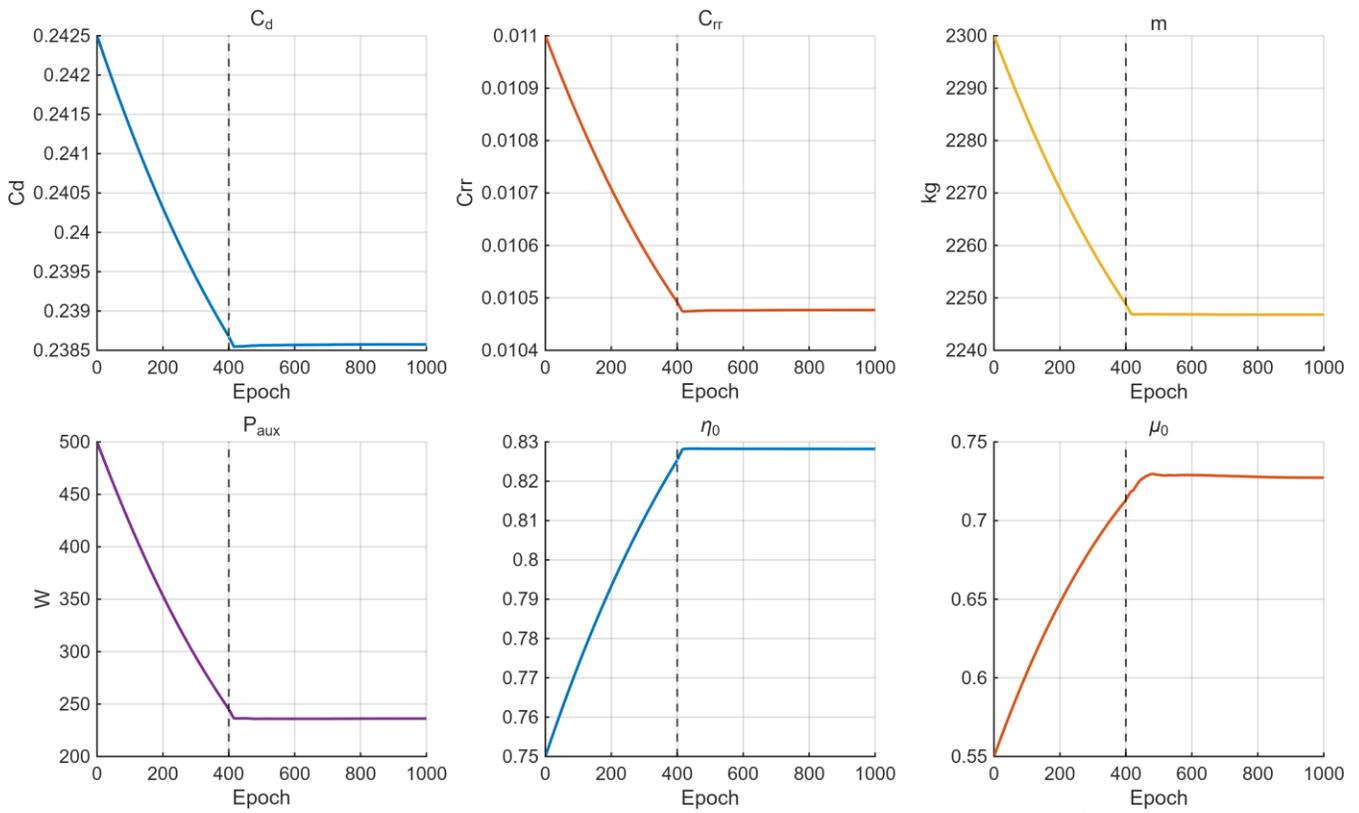

**Fig. A-2.** Model S parameter evolution. Broad priors contract to plausible, bounded values; clipping fades late in training.

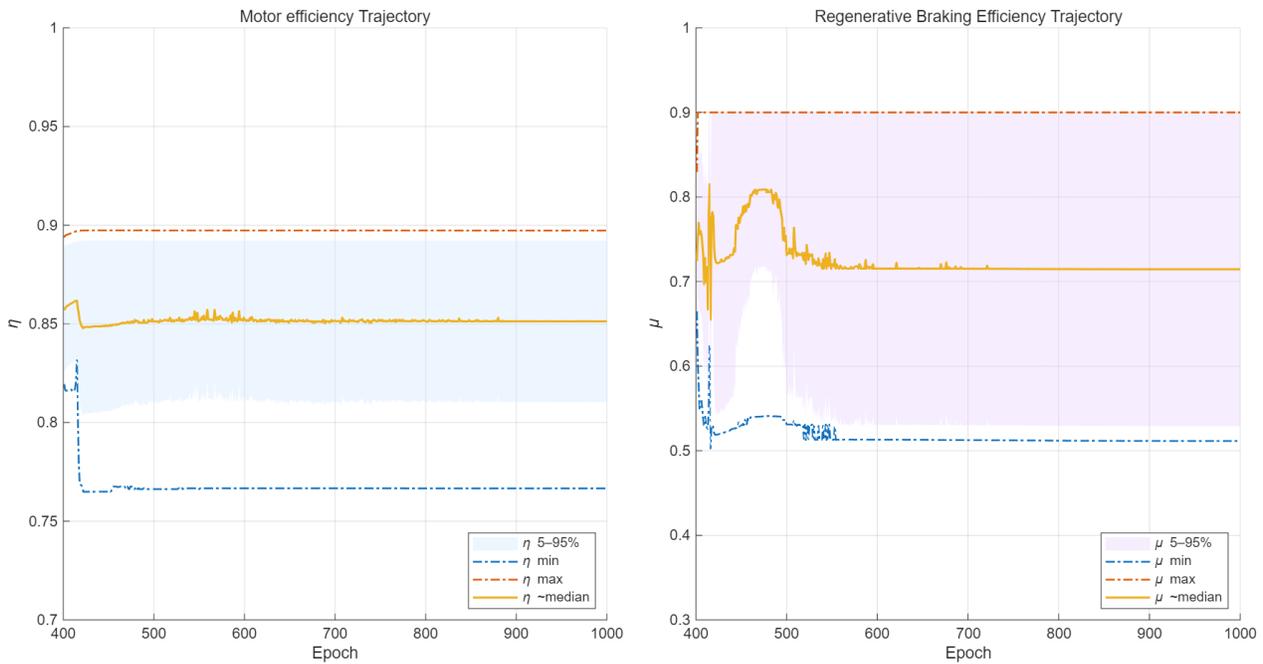

**Fig. A-3.** Model S efficiencies. Motor efficiency tracks load; Regenerative braking efficiency activates during deceleration only.

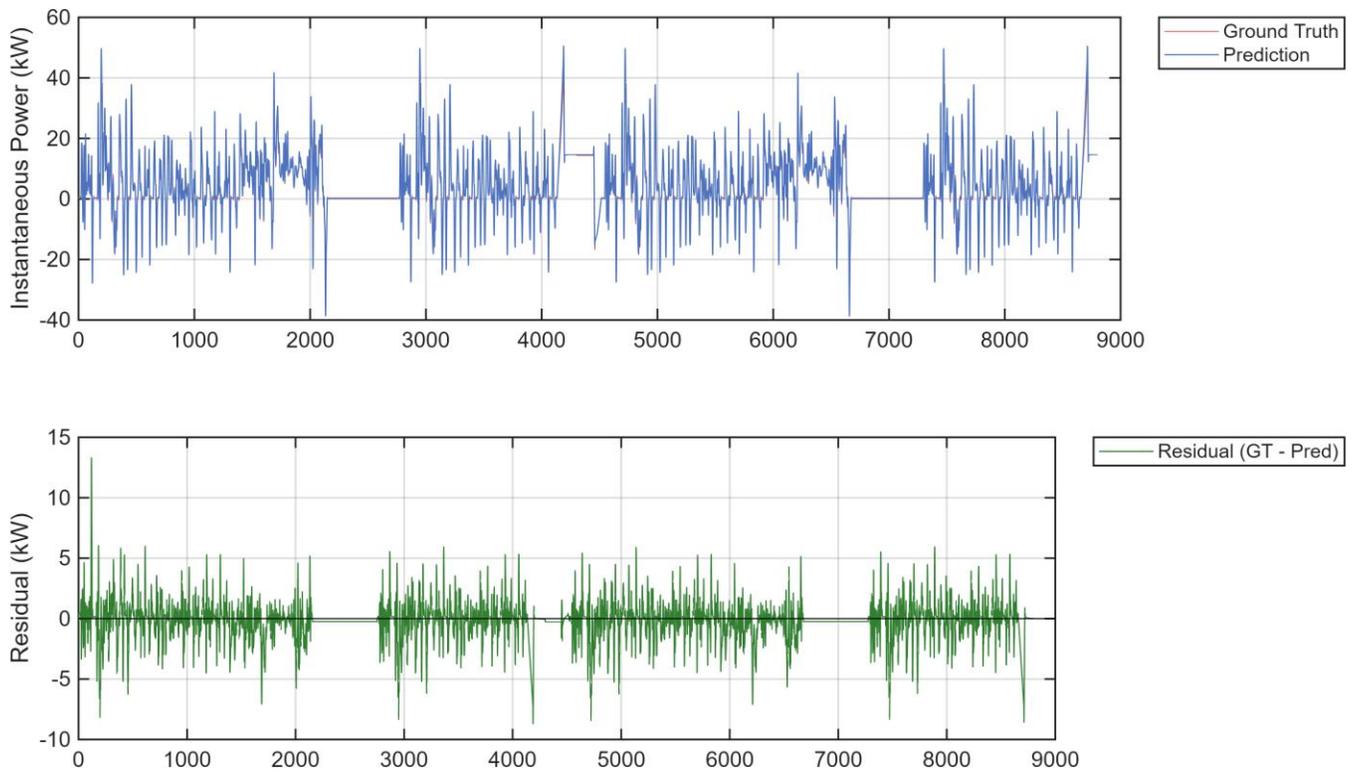
**Fig. A-4.** Model S prediction vs. ground truth. Tight overlap with near-zero-mean residuals and no visible lag. Low MAE and RMSE at ~0.19 kW and ~0.29 kW, respectively.

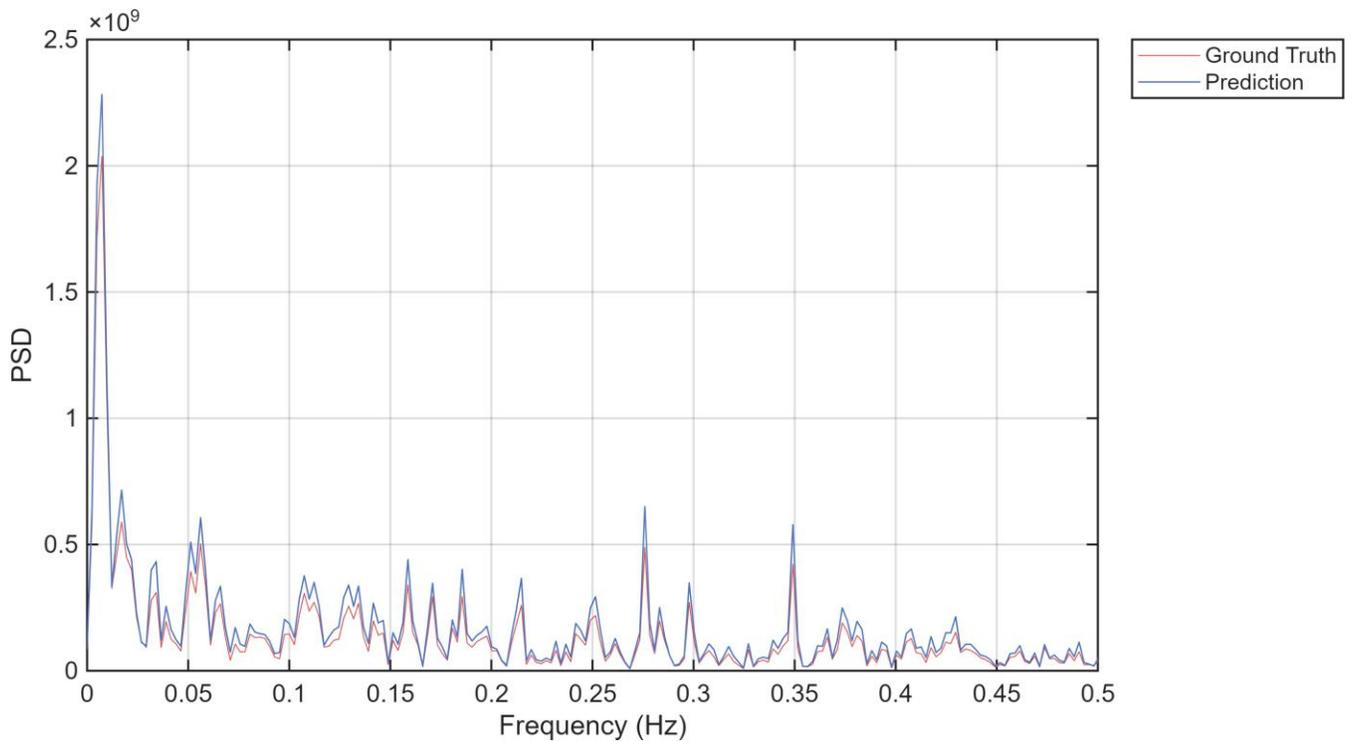
**Fig. A-5.** Model S PSD test. Matched spectral peaks and noise floor, indicating correct dynamic content without artifacts.

**Kia EV9 Long Range 4WD**

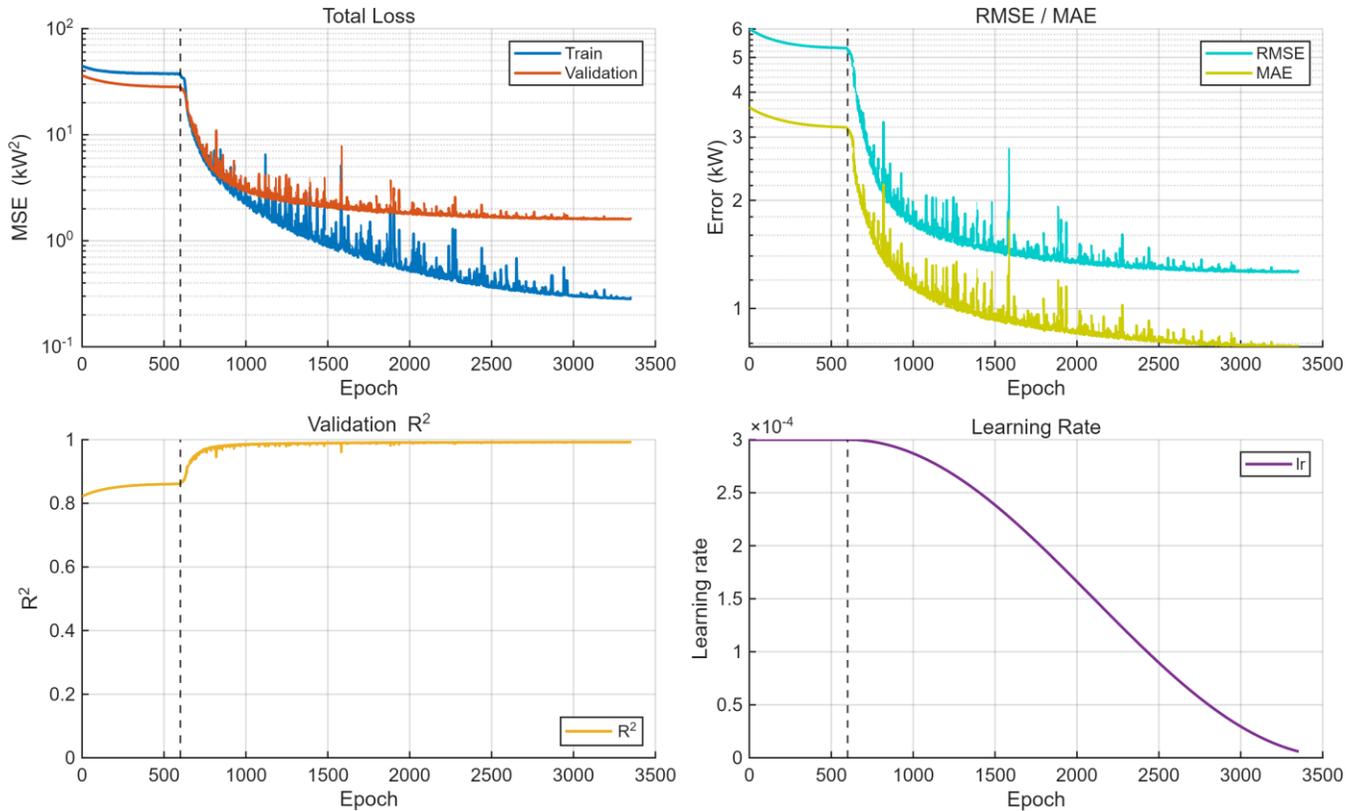

**Fig. A-6.** EV9 training curves. The EV9's training curves were relatively less stable, which we attribute to its low-resolution speed data. It was taken from dashboard display memory, which has quantization error from rounding to nearest integer. Nevertheless, the model achieved a low MAE of 0.78 kW and RMSE of 1.26 kW, which is very good considering the EV9's power range is roughly five times larger than the Tesla datasets. The observed peak power of ~280 kW is consistent with the vehicle's official 283 kW (380 hp) specification and suggests the learned model can accurately capture highly dynamic highway accelerations, even with a challenging dataset.

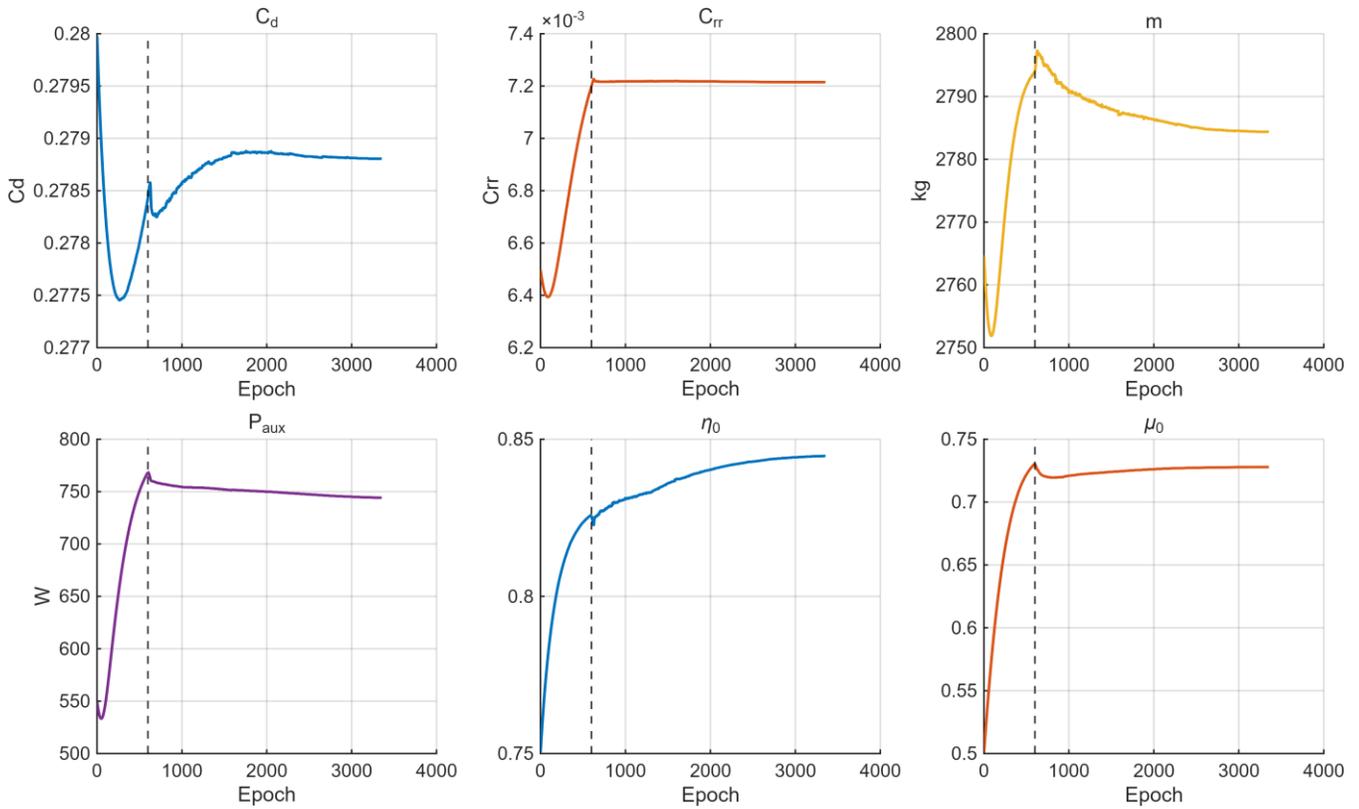

**Fig. A-7.** EV9 parameter evolution. The values settle to SUV-appropriate values.

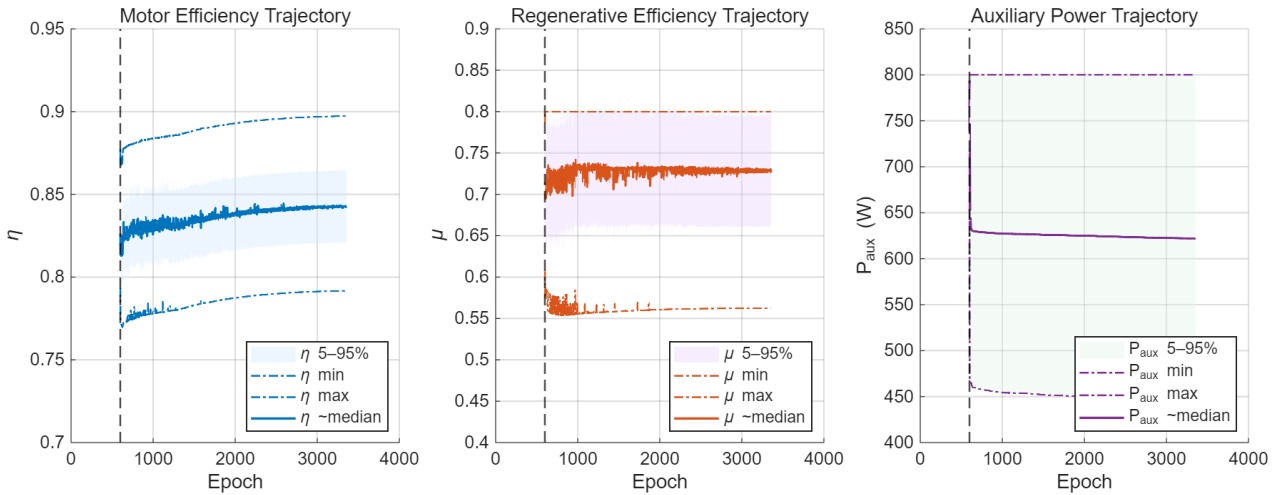

**Fig. A-8.** EV9 efficiency trajectories. The trajectories follow the operating regime. We also model auxiliary power as variable head for this case. We noticed that the auxiliary power for EV9 fluctuates hugely which implies that the passengers changed settings in HVAC/lights/accessories.

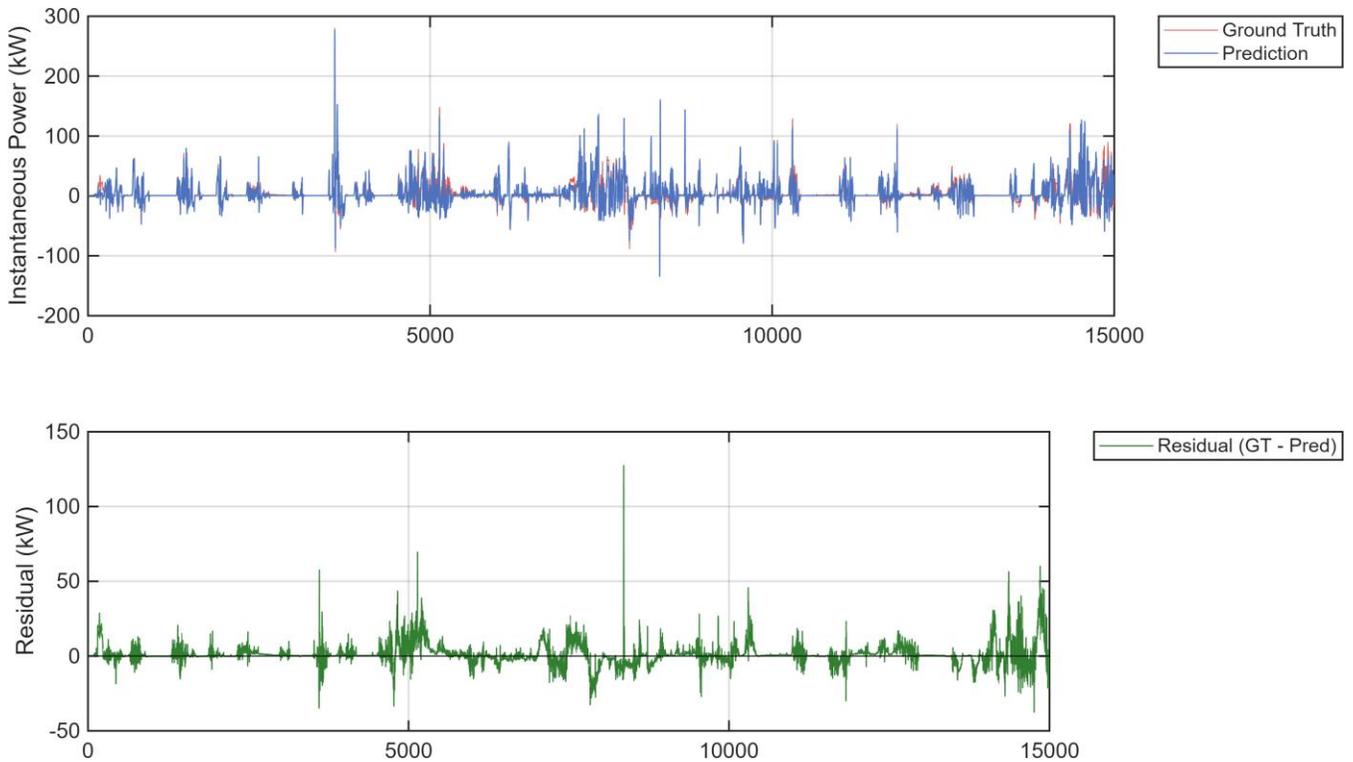

**Fig. A-9.** EV9 prediction vs. ground truth with variable $P_{aux}$. Residuals (~0.5 kW) largely arise from $P_{aux}$, because the mapping from velocity and acceleration to $P_{aux}$ is not represented. Since $P_{aux}$ is human-triggered, median values are preferable for inference. We can also consider that when $P_{aux}$ is set as variable parameter, it may soak up unmodeled forces and unlike residual buffer head, $P_{aux}$ is not regularized to be small. This can be fixed, however, but we believe it is best to leave $P_{aux}$ as static variable and instead learn the median value that best fits the data.

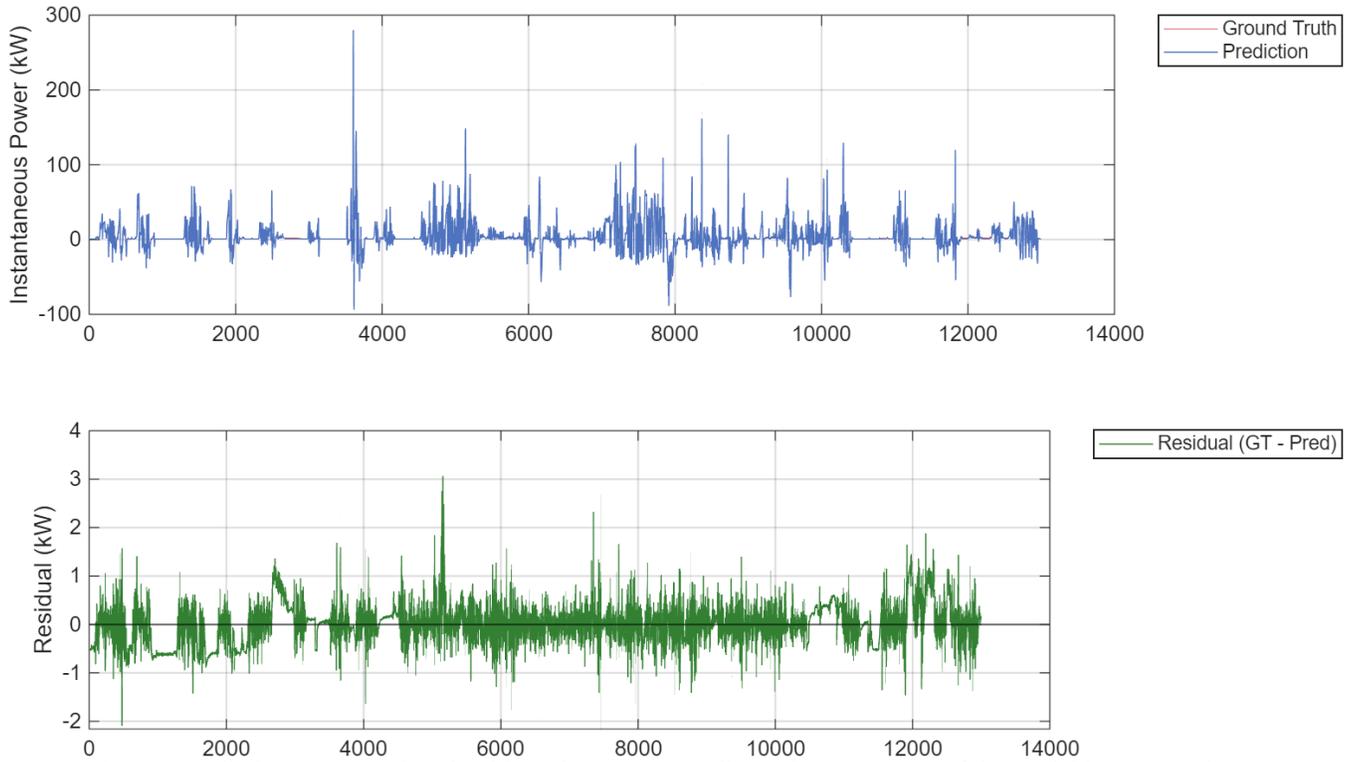

**Fig. A-10.** EV9 prediction vs. ground truth with static $P_{aux}$ at median value. Accuracy of the model has restored.

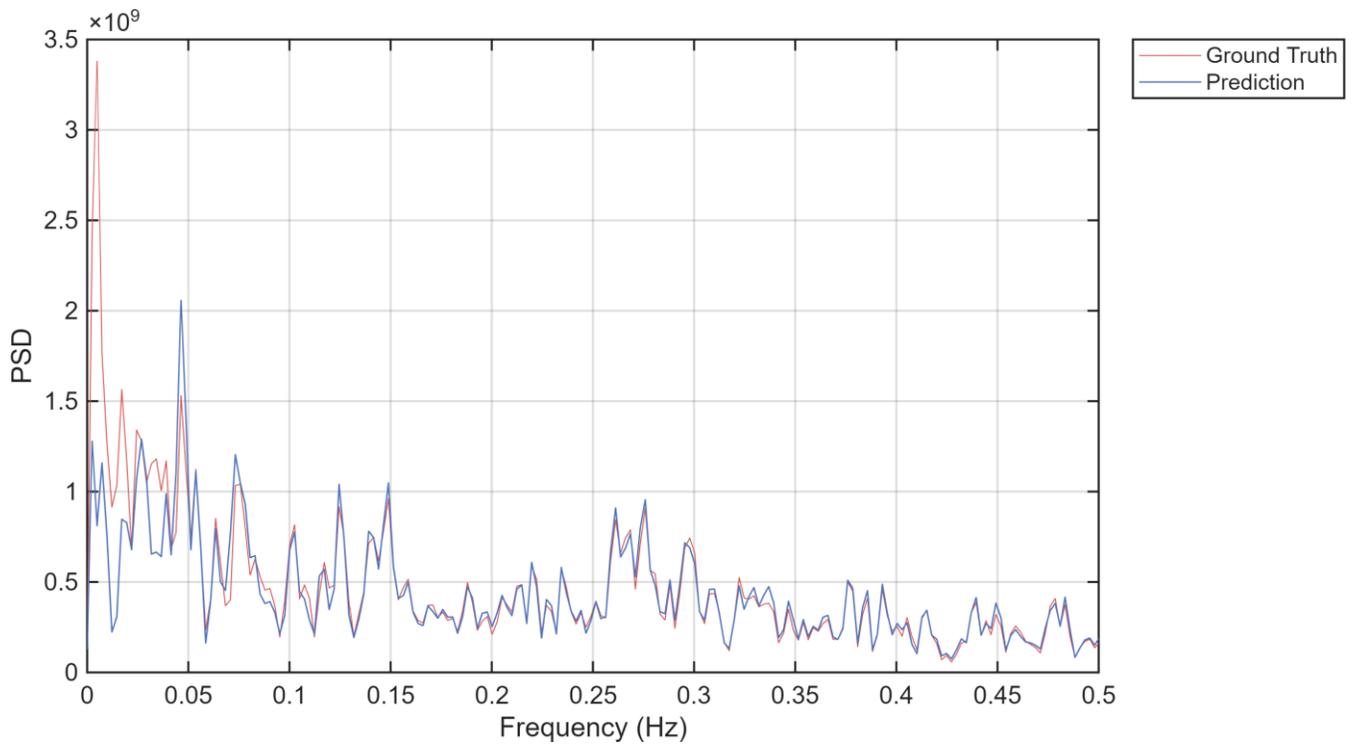

**Fig. A-11.** EV9 PSD test with variable $P_{aux}$: Since $P_{aux}$ varies only at very low frequencies typically when someone adjusts HVAC or toggles other auxiliary loads. Otherwise, it acts as a baseline draw across the entire drive, so even small power levels integrate to substantial energy over time.

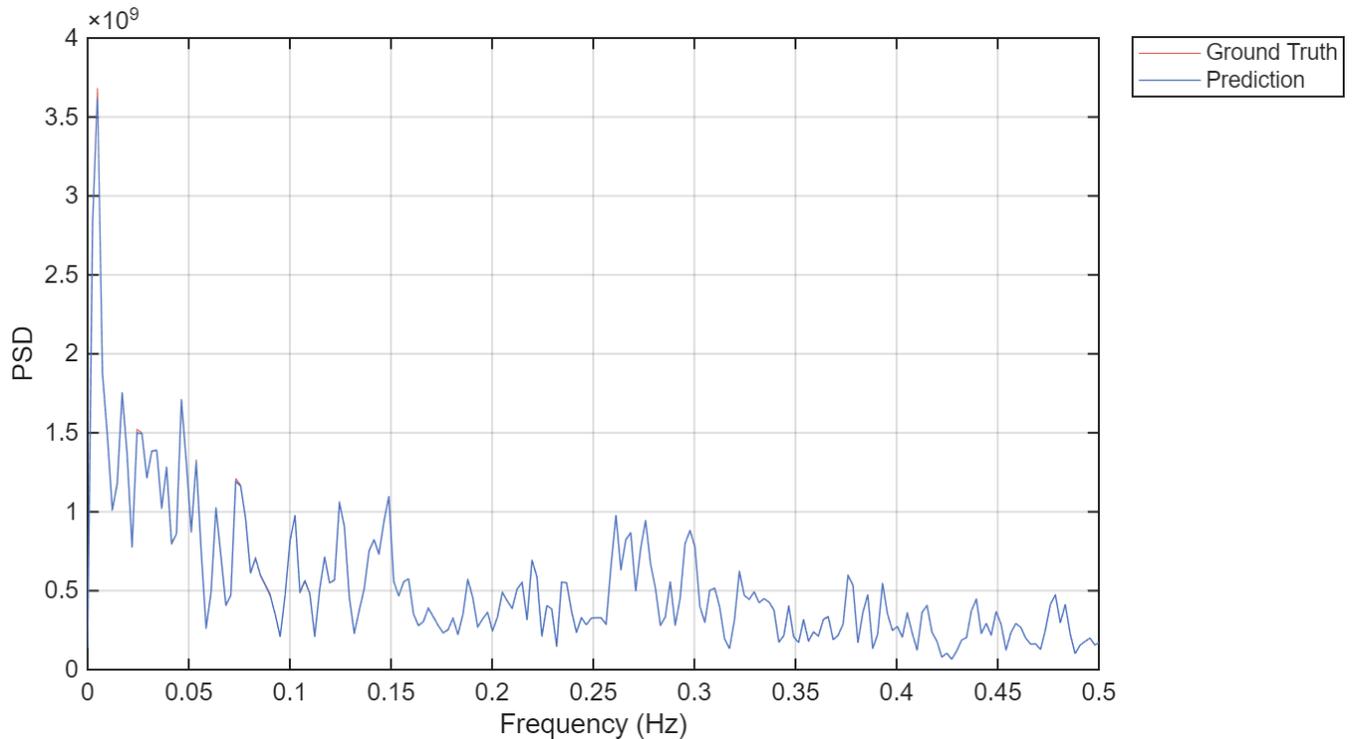

**Fig. A-12.** EV9 PSD test with static $P_{aux}$ at median value: dominant peaks and noise floor align with ground truth, confirming frequency-domain fidelity. We hypothesize that the low-pass filtering effect of using a limited number of Fourier modes ($m = 4$ in this case) was coincidentally well-suited to regularize this specific type of noise.

## APPENDIX B

The hyperparameters used for training EV-PINO models are summarized in Table B-1.

| Parameter | Value / Setting |
|---|---|
| Fourier modes ($m$) | 4 |
| Spectral Layers | 4 layers, width 128 |
| Activation function | GELU |
| Sequence window length ($L$) | 128 |
| Stride | 32 |
| Batch size | 128 |
| Optimizer | Adam ($lr = 3 \times 10^{-4}$) |
| LR scheduler | Cosine annealing, $lr_{min} = 10^{-6}$ |
| Warm up epochs | 400 (3/S) / 600 (EV9) |
| Max epochs | 3500 |
| Early-stopping patience | 200 epochs |

**Table B-1.** Hyperparameters used for EV-PINO training and validation. Same hyperparameters were kept for training the models for all vehicles. The Kia EV9 model was given 600 warm-up epochs.

### *Choosing the number of Fourier modes*

In the FNO trunk, the spectral blocks only learn on the lowest frequency $m$ rFFT coefficients. Too few modes will underfit slow power trends. Having too many modes wastes compute and can overfit high-frequency noise.

In our case, we sampled 10Hz sensor sample then applied an 11-point order 3 SG filter.

$$f_c \approx \frac{f_s}{W} \approx \frac{10}{11} \approx 0.91 \; Hz \tag{B-1}$$

$$\Delta f = \frac{f_s}{L} = \frac{10}{128} = 0.078125 \; Hz \tag{B-2}$$

where $W$ is the SG window length in samples.

To ensure the spectral trunk can represent all energy left after filtering, the highest kept bin $(m-1)\Delta f$ should reach $f_c$:

$$m \geq \frac{f_c}{\Delta f} + 1 \Rightarrow m \gtrsim \frac{0.91}{0.078125} + 1 \approx 12.6. \tag{B-3}$$

Thus, a conservative upper bound for $m = 13$.

However, we only used 4 Fourier modes because 95% of the EV energy is low-frequency. In practice, most variance sits well below 0.5Hz (as shown in Figs. A-5 and A-12). Having smaller $m$ also reduces spectral search space and removes extra computation that only makes the network to chase for noise.